%% file: WAFR16.tex
\documentclass{llncs}

\usepackage{algorithm}
\usepackage{algorithmic}
\usepackage{amsfonts,amssymb}
\usepackage{dsfont}
\usepackage{graphicx}
\usepackage{color}
\usepackage{xcolor}
\usepackage[font=small,labelfont=bf]{caption}
\usepackage{subcaption}
\usepackage{tikz}
\usepackage{url}

\begin{document}
\title{Estimating Activity at Multiple Scales \\using Spatial Abstractions}

\author{Majd Hawasly \inst{1} \and Florian T. Pokorny\inst{2} \and Subramanian Ramamoorthy\inst{1}}

\institute{School of Informatics, University of Edinburgh \and KTH Royal Institute of Technology}

\maketitle
\begin{abstract}
Autonomous robots operating in dynamic environments must maintain beliefs over a hypothesis space that is rich enough to represent the activities of interest at different scales. This is important both in order to accommodate the availability of evidence at varying degrees of coarseness, such as when interpreting and assimilating natural instructions, but also in order to make subsequent reactive planning more efficient. We present an algorithm that combines a topology-based trajectory clustering procedure that generates hierarchically-structured spatial abstractions with a bank of particle filters at each of these abstraction levels so as to produce probability estimates over an agent's navigation activity that is kept consistent across the hierarchy. We study the performance of the proposed method using a synthetic trajectory dataset in 2D, as well as a dataset taken from AIS-based tracking of ships in an extended harbour area. We show that, in comparison to a baseline which is a particle filter that estimates activity without exploiting such structure, our method achieves a better normalised error in predicting the trajectory as well as better time to convergence to a true class when compared against ground truth.
\end{abstract}

\section{Introduction}
\input{Ram_introduction.tex}

\section{Hierarchical abstraction of trajectories}
To create the filtration of spatial abstractions from trajectories, we consider hierarchical clustering~\cite{clustering} by means of discrete Fr\'{e}chet distance~\cite{discrete_frechet}. {\color{black}For two discretised $d$-dimensional trajectories $\tau_1:[0,m]\rightarrow\mathds{R}^d$ and $\tau_2:[0,n]\rightarrow\mathds{R}^d$: 
\begin{equation}
\delta_F(\tau_1,\tau_2)=\inf_{\alpha,\beta}  \max_{j\leq m+n} \delta_E(\tau_1(\alpha(j)),\tau_2(\beta(j))),
\end{equation}
 where $\alpha$ and $\beta$ are discrete, monotonic re-parametrisations $\alpha:[1:m+n]\rightarrow[0:m],\;\beta:[1:m+n]\rightarrow[0:n]$ which align the trajectories to each other point-wise, and $\delta_E(.,.)$
 is the Euclidean distance between two points.} This metric $\delta_F$ corresponds to the maximal point-wise distance between two optimal
 reparameterisations of $\tau_1, \tau_2$, and it can be computed efficiently using dynamic programming in $O(mn)$ time~\cite{discrete_frechet}. 

\begin{figure}[th]
\centering
 \def\svgwidth{0.99\textwidth}
{ 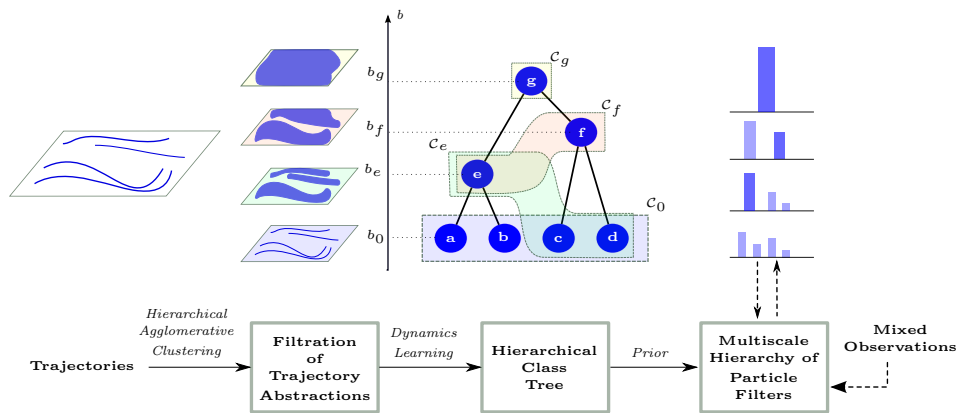}
\caption{\footnotesize An overview of the approach. A collection of trajectories are clustered at different resolutions to give a filtration of spatial abstractions, and dynamics  models are acquired for the individual classes. The classes are organised in a tree structure with their birth indices. The shaded areas show levels of the hierarchy, where $\mathcal{C}_0$ is the finest level with single trajectory classes, and $\mathcal{C}_g$ is the coarsest class, representing all the trajectory data combined. The distributions on the right show an example of a  consistent distribution across the tree. A collection of particle filters encode these distributions and allow coarse observations to be handled at any level of the hierarchy.}
\label{fig:overview}
\end{figure}

After computing the distance matrix $D$ for the trajectories, where $D_{i,j}=D_{j,i}=\delta_F(\tau_i,\tau_j)$ and
$D_{i,i}=0$, the trajectories can be considered as data points to be clustered, $C=\{\tau_1,\tau_2,\ldots,\tau_M\}$. A hierarchical agglomerative clustering~\cite{hierarchical_clust} of $C$ results in a tree of trajectory clusters (see Figure~\ref{fig:overview}).

Hierarchical agglomerative clustering is an iterative approach to data clustering in which, at every iteration, two clusters at a lower level get merged to make a single new cluster. This gives rise to a tree data structure in which the leaves at one end are the individual data items, and the root node at the other end is the cluster made by merging all data points together, while the intermediate layers combine data items based on their proximity. The order of merging depends on the distance between the clusters, such that the pair with the smallest distance are merged first. Thus, every new cluster can be assigned a distance value at which it gets created. An important consideration in the design of hierarchical agglomerative clustering algorithms is the method of computing  the distance between clusters. 

A \emph{single linkage} algorithm assumes the distance between two  clusters to be the smallest distance between the individual data items in the two clusters. At stage $t$, for a collection of objects $C^t$ and a distance matrix $D^t$, the pair of distinct clusters with the smallest distance in $D^t$ are merged  to create a new cluster,  $\tau_{ij}=\cup_{h} \tau_h\,\,$ for $\; h \in\arg\min_{i,j, i\neq j} D^t_{i,j} $.  
$C^{t+1}$ is created by removing the two clusters $\tau_i$ and $\tau_j$ and adding $\tau_{ij}$. The distance matrix is also updated to reflect the change, 
$ D^{t+1}_{ij,h} = \min(D^t_{i,h},D^t_{j,h})$ for all $\tau_h\in{C^t}$. The process repeats until it terminates when only one cluster remains.

We call $D_{i,j}$ the \emph{birth index} of cluster $\tau_{ij}$, and we denote it by $b_{ij}$. This refers to the distance threshold after which $\tau_{ij}$ starts to exist.   Also, we define the \emph{death index} $d_i=d_j=D_{i,j}$ to be the distance index at which cluster $\tau_i$ (similarly, $\tau_j$) ceases to exist.

The output of this algorithm is a tree structure $\mathcal{T} \langle \mathcal{C},\rho\rangle$ where
$\mathcal{C}=\bigcup_t C^t$ is a  collection of all the hierarchical clusters (original ones included), and the parent function $\rho:\mathcal{C}\rightarrow
\mathcal{C}$ which maps a cluster to its immediate \emph{parent}.  Then, if $\rho(\tau_i)=\tau_h$, then $\tau_i\subset \tau_h$ and $b_h=d_i$; see the tree
in Figure~\ref{fig:overview}.

We consider a tree node $\tau_i\in\mathcal{C}$ to be \emph{alive} at some birth index $b$ if $b_i\leq b<d_i$, i.e., when it is born but not yet dead. A \emph{level} in the tree, identified with a birth value $b$, contains all the nodes that are
 alive at $b$. We denote the level at index $b$ by $\mathcal{C}_b\subseteq\mathcal{C}$.

\begin{figure}[th!]
\centering
\begin{subfigure}{0.8\textwidth}
    \fcolorbox{lightgray}{white}{\includegraphics[width=.20\textwidth]{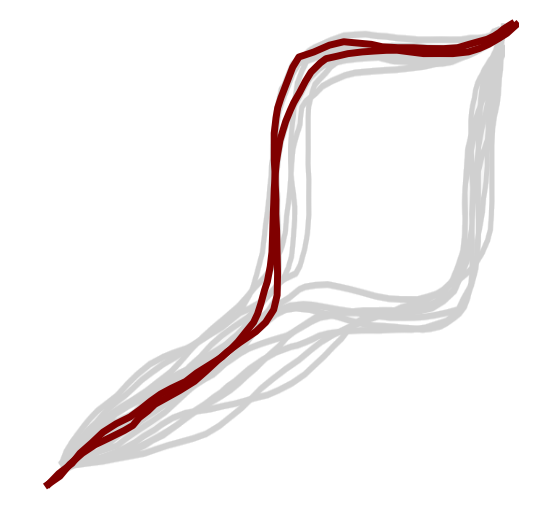}}  \fcolorbox{lightgray}{white}{\includegraphics[width=.20\textwidth]{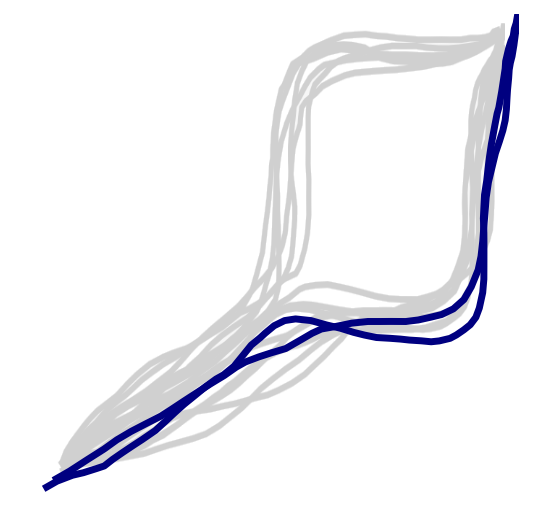}}  \fcolorbox{lightgray}{white}{\includegraphics[width=.20\textwidth]{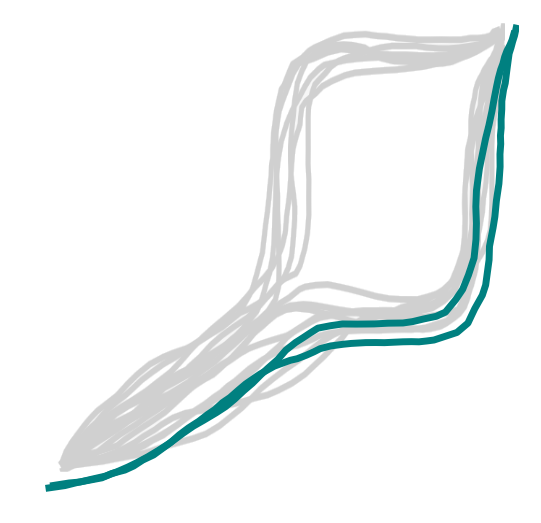}} \fcolorbox{lightgray}{white}{\includegraphics[width=.20\textwidth]{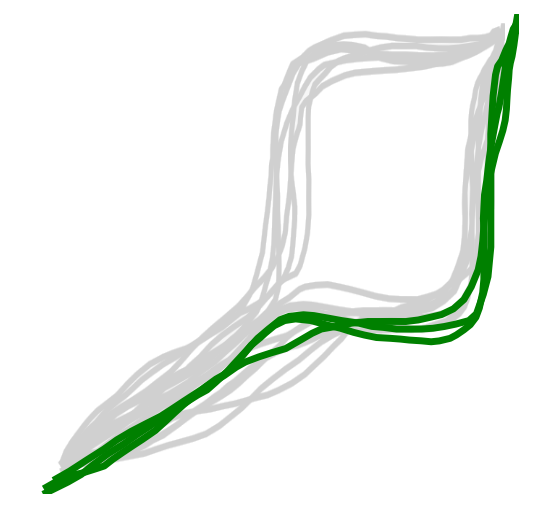}}
\end{subfigure}%
\\
\begin{subfigure}{0.8\textwidth}
    \fcolorbox{lightgray}{white}{\includegraphics[width=.20\textwidth]{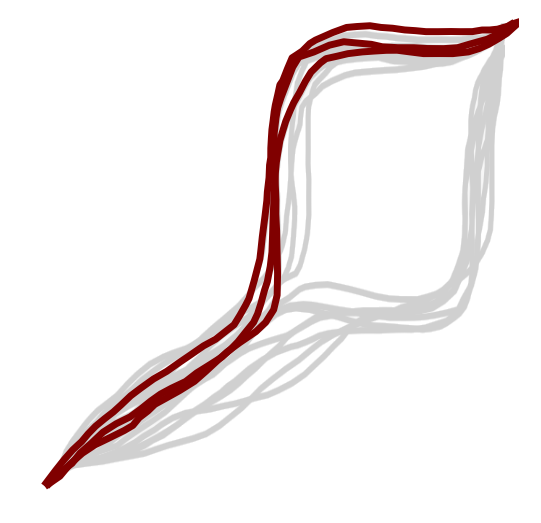}}  \fcolorbox{lightgray}{white}{\includegraphics[width=.20\textwidth]{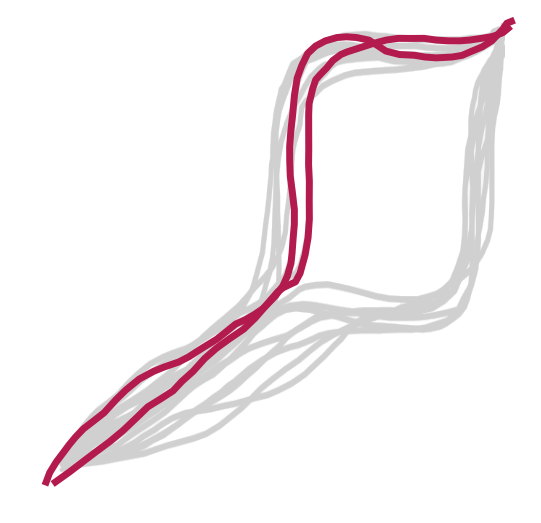}}  \fcolorbox{lightgray}{white}{\includegraphics[width=.20\textwidth]{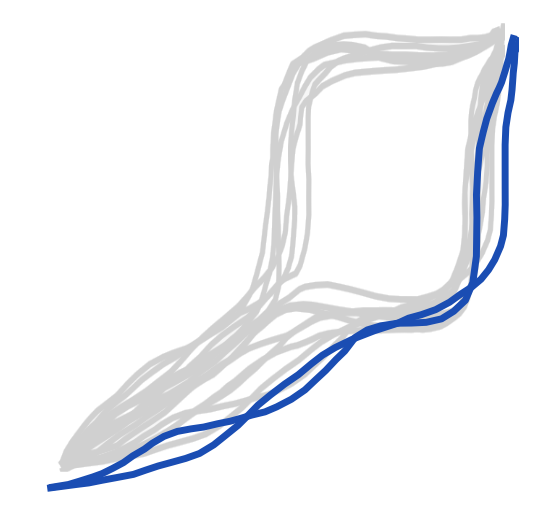}} \fcolorbox{lightgray}{white}{\includegraphics[width=.20\textwidth]{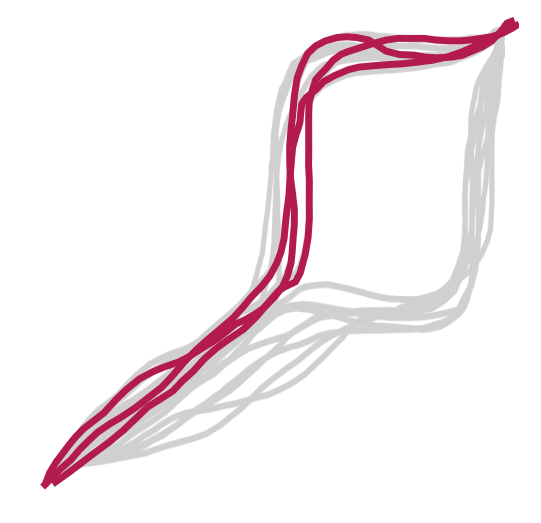}} 
\end{subfigure}%
\\
\begin{subfigure}{0.8\textwidth}
    \fcolorbox{lightgray}{white}{\includegraphics[width=.20\textwidth]{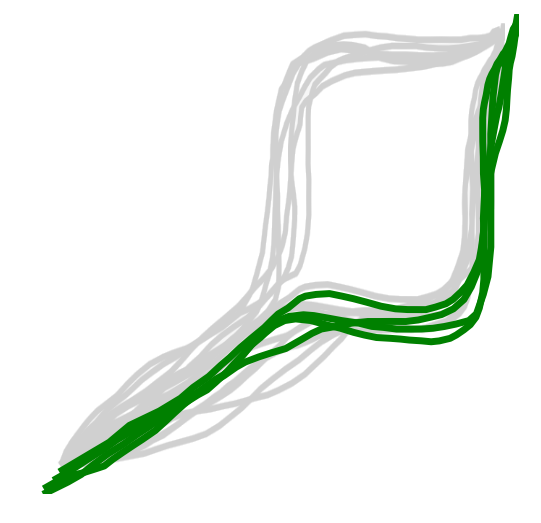}}  \fcolorbox{lightgray}{white}{\includegraphics[width=.20\textwidth]{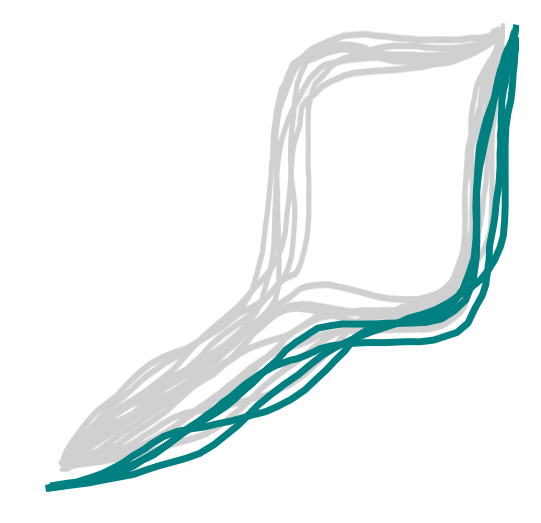}}
\fcolorbox{lightgray}{white}{\includegraphics[width=.20\textwidth]{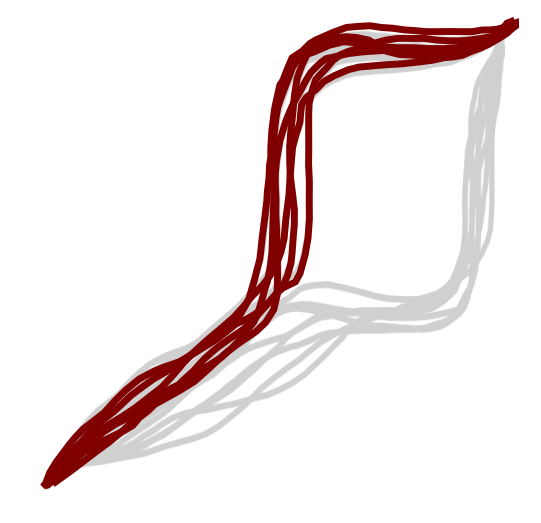}}  \fcolorbox{lightgray}{white}{\includegraphics[width=.20\textwidth]{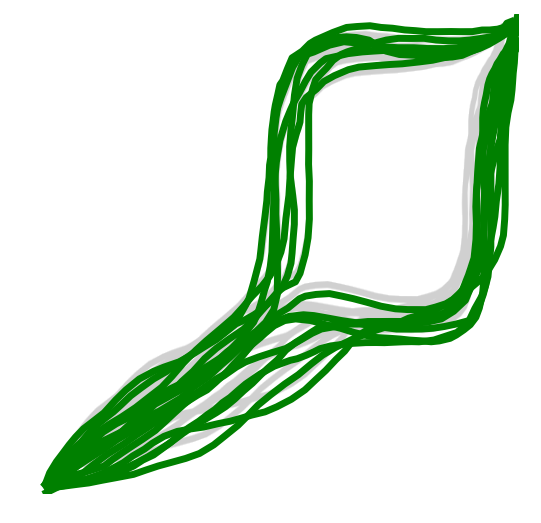}}
\end{subfigure}%
\caption{\footnotesize Clustering a set of 14 trajectories using hierarchical single-linkage agglomerative clustering with Fr\'{e}chet distance. Panels show the new clusters at different levels of the filtration (tree levels). Note that the rest of the trajectories/clusters at each level are depicted in the background for the sake of clarity.}
\label{fig:trajs}
\end{figure}

Figure~\ref{fig:trajs} illustrates clustering 14 representative trajectories of navigation around a Y-junction using the method described above. Each panel shows the newly-created cluster in one level of the hierarchy at some birth index. The rest of the trajectories appear in grey in the background.
%

\section{\color{black}Multiscale Hierarchy of Particle Filters\label{sec:MHPF}}
A Bayesian particle filter~\cite{doucet2009tutorial} tracks a probability distribution (a \emph{belief}) over some random variable of interest $z^t\in\mathds{R}^d$ by evolving a collection of $N$ hypotheses  called \emph{particles} utilising  prior knowledge $\mathbf{P}(z^0)$ and a sequence of  measurements $\xi^{1:t}$. 

Upon receiving a new observation $\xi^t$, a Bayesian belief should be updated as follows: \[\mathbf{P}(z^t|\xi^{1:t})\propto\mathbf{P}(\xi^t|z^t)\sum_{z^{t-1}\in\mathds{R}^d} \mathbf{P}(z^t|z^{t-1})\mathbf{P}(z^{t-1}|\xi^{1:t-1}).\] Sampling directly from the target distribution might not be feasible, so a particle filter computes an approximation that involves representing its beliefs by a set of particles. In the standard algorithm~\cite{probabilistic_robotics}, particles are sampled from a \emph{proposal distribution} (typically, the dynamics model $\mathbf{P}(z^t|z^{t-1})$) and the deficit between the two distributions is rectified by assigning \emph{importance weights} $w^t\in[0,1]$ to the particles (typically, the observation likelihood $\mathbf{P}(\xi^t|z^{t})$). The actual belief is retained again by \emph{resampling} $N$ particles according to their weights. In practice, one replaces a small fraction of all particles randomly with new ones regardless of their weights in a bid to avoid particle \emph{depletion}.

We present the Multiscale Hierarchy of Particle Filters (MHPF), a stack of \emph{consistent} particle filters defined over \emph{abstractions} of the value of the random variable. {\color{black}In this paper, the random variable of interest is the agent's navigation plan, encoded quantitatively in its point position, and qualitatively in the class and shape of its planned trajectory. We assume that a collection of trajectories are available to MHPF in order to construct this abstraction from data.} The abstractions are representations of this random variable at decreasing resolution, so that the lowest level of the abstractions hierarchy consists of complete trajectories at the smallest scale (with cardinality equal to the size of the trajectory dataset). At any higher level, these trajectories are clustered into a  smaller number of bins or categories, representing coarser descriptions of the trajectory shapes. Thus, at any stage the status of the process can be queried at any of the different levels of resolution. This gives the additional advantage of allowing evidence at various degrees of coarseness to be incorporated into the filter by using it to update the probability estimate at that level. In order to maintain consistency between the particle filters of the stack, this update results in corresponding updates to all the other filters - based on a procedure to be described below.

\subsection{Setup}
Given a tree $\mathcal{T} \langle \mathcal{C}, \rho\rangle$ over the input trajectories, each cluster $c\in\mathcal{C}$ is a collection of `similar' trajectories  at some level of resolution $b$. The cluster $c$ then can be seen as a \emph{class} of behaviour for the tracked process, identified by a generative dynamics model $\theta=\mathbf{P}(z'|z,c)$, $z,z'\in\mathds{R}^d$ from which the member trajectories are samples. 

The dynamics of a class $c$ are approximated from the points of the member trajectories using a localised model as follows. All the points of   $c$ in an $\epsilon$-ball around the point of interest $z\in\mathds{R}^d$ are located $B_\epsilon(z)=\{z'\in c:
\delta_E(z',z)< \epsilon\}$, and the local velocities at these points $\dot{z}'$ are used to estimate the new velocity, $\dot{z}=\eta\,\sum_{{z'\in B_\epsilon(z)}}{1/\delta_E(z,z')\,\dot{z}'}$, where $\eta^{-1}={\sum_{z'\in B_\epsilon(z)}{1/\delta_E(z,z')}}$ is a normalisation factor. Then, a new position is sampled, $z^*\sim z+\dot{z}+\gamma(\kappa)$, where $\gamma(\kappa)$ is a noise term related to dynamics noise parameter $\kappa$.\label{localised_dynamics} 

The tree of clusters $\mathcal{T}$ and the associated dynamics are the input to MHPF.  As a stack of filters, a distinct filter is defined for every level of the tree $\mathcal{T}$. Thus, a particle $x^t$ in MHPF is a weighted hypothesis of the \emph{class and the position} at time $t$. That is, every particle at some level $b$ represents a hypothesis  not only for the position of the tracked process in $\mathds{R}^d$ but also which of the different classes in $\mathcal{C}_b$ represents the behaviour of the process best. We write $(x^t(z^t,c),w^t)$ where $z^t\in\mathds{R}^d$ is the position, $c\in\mathcal{C}_b$ is the class, and $w^t$ is a weight that reflects to what extent the hypothesis of the particle is compatible with the evidence. We denote by $X_b$ the set of particles at level $b$.

There are two kinds of observations in MHPF: 1)~the noisy position observations $z^t+\gamma(\psi)$, where $\gamma(\psi)$ is a noise term related to the observation noise parameter $\psi$, which are the typical observations for standard particle filters as well as the filter of the lowest level of the MHPF stack; and 2)~\emph{coarse observations} which provide evidence regarding the underlying process and can be identified to one of the classes in $\mathcal{C}$ other than $\mathcal{C}_0\;$\footnote[1]{\color{black}This is compatible both with variable resolution sensors (e.g. GPS receivers) and with high-level qualitative instructions (e.g. linguistic instructions) as long as a mapping can be established between the observation and $\mathcal{C}$, especially in the latter type.}. In both cases, the MHPF returns a stack of consistent probability distributions pertaining to the different tree levels.

\subsection{Procedure}
MHPF is based on a  probability distribution defined at the finest level $\mathcal{C}_0$ from which the tree is  rebuilt, as shown in the procedure in Algorithm~\ref{alg_mpf}.

First, the particle set $X_0$ is created by sampling $N$ particles from a prior $\Delta(\mathcal{C}_0\times \mathds{R}^d)$ over class assignment and initial positions, then assigning them equal weights, where $\mathcal{C}_0=\{c\in\mathcal{C}\, |\, b_c=b_0\}$ is the collection of individual trajectories forming a class each at the lowest clustering threshold $b_0$. Denote by $N_i$ the number of particles from class $c_i$, such that $\sum_{c_i\in\mathcal{C}_0}
N_i=N$.


\begin{algorithm}[th!]
 \caption{Multiscale Hierarchy of Particle Filters}
 \label{alg}
 \begin{algorithmic}[1]
\REQUIRE Prior over particles, number of basic particles $N$, the depletion parameter $v$, tree structure $\mathcal{T}$    

\STATE Sample $N$ particles from the prior $\Delta(\mathcal{C}_0\times \mathds{R}^d)$ with  equal weights.
\FOR{each time step $t>0$}{
\STATE Build the tree probabilities up from $\mathcal{C}_0$ (Algorithm~\ref{alg:tree_rebuild}).
\FOR{parents $\bar{c}$ of  ${\mathcal{C}}_0$ recursively \TO the top of $\mathcal{T}$}
\STATE Sample a number of particles from $\bar{c}$; ${N}_{\bar{c}}=\sum_{\underline{c}=\rho^{-1}(\bar{c})} N_{\underline{c}}$, with equal weights. 
\ENDFOR
\FOR{each  particle $x(z^{t-1},c)$}
{
\STATE sample a new position $z^t\sim \mathbf{P}(z|z^{t-1},c)$
}
\ENDFOR
\STATE Receive observation $\xi^t$.
\STATE Update tree with observation $\xi^t$ (Algorithm~\ref{alg:obs_update}).
\STATE  Resample $N\,(1-v)$ particles from $\mathcal{C}_0$ based on the updated weights $w^t$. 
\STATE  Sample $N\,v$ particles uniformly randomly from $\mathcal{C}_0$.
}\ENDFOR
 \end{algorithmic}
\label{alg_mpf}
\end{algorithm}


The probabilities of the classes of $\mathcal{C}_0$ are computed from the initial weights, and these probabilities in turn are used to compute the probabilities of the rest of the classes as described in Algorithm~\ref{alg:tree_rebuild}. 

\begin{figure}[ht!]
\centering
\begin{subfigure}{0.40\textwidth}  
\centering
    \includegraphics[width=.55\textwidth]{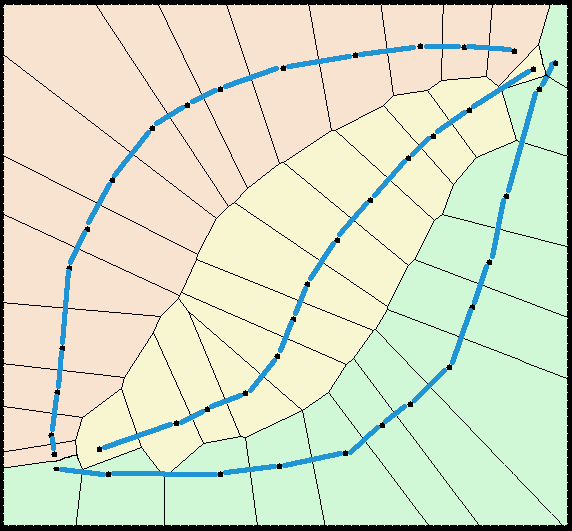} 
  \caption{\footnotesize Three separate classes}
\end{subfigure}%
\begin{subfigure}{0.40\textwidth}  
\centering
    \includegraphics[width=.55\textwidth]{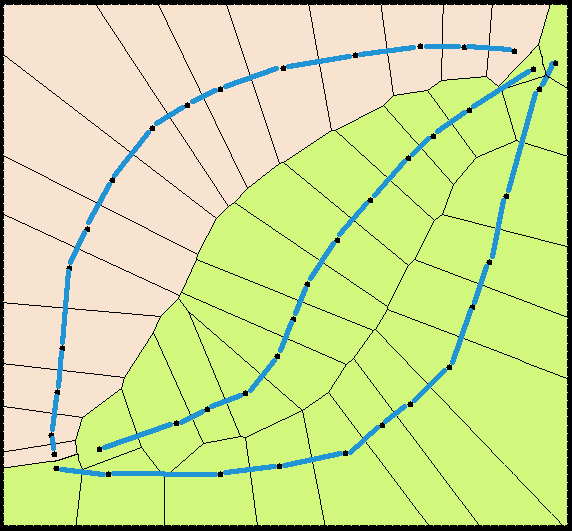} 
\caption{\footnotesize Two classes merge}
\end{subfigure}%
\caption{\footnotesize An illustration of the intuition behind the relationship between the spatial abstraction and the corresponding probability operations on the tree in a toy example 2-dimensional domain. \label{delaunay}}
\end{figure}

At this stage, the class probabilities are propagated recursively upwards by the additivity rule, so that a parent's probability is the sum of its children's probabilities, $\mathbf{P}^t(\bar{c})=\sum_{\underline{c}=\rho^{-1}(\bar{c})} \mathbf{P}^t(\underline{c})$. In order to understand the intuition behind this step, consider the probabilities assigned to the classes/nodes of the tree with respect to the \emph{regions} that are defined by  a spatial nearest neighbour relationship to the points of their corresponding trajectories. Consider the example of a 2-dimensional domain in Figure~\ref{delaunay}, where the region corresponding to a class can be understood as  the union of 2-dimensional Voronoi cells of a discretisation of the class trajectories. Thus, merging two classes in the tree is analogous to merging the regions associated with their classes, and correspondingly adding the probability of the two child classes to yield the probability of the parent. Similarly, the children of a class proportionally inherit their parent's probability when moving downward in the tree. The Voronoi cells depicted below are never explicitly computed. However, implicitly, this defines our notion of consistency between the probability estimates at the levels of the hierarchy. 


With the probabilities specified, the same number of particles as the total number of children's particles are created for the parent $\bar{c}$ ,
$N_{\bar{c}}=\sum_{\underline{c}=\rho^{-1}(\bar{c})} N_{\underline{c}}$, and this is repeated recursively to the top of the tree.  Note that any arbitrary level $b$ of the tree would have exactly $N$ particles with a proper probability distribution, while the total number of particles in the full tree depends on how the particles are distributed between the tree branches. The last stage of the tree construction is to sample new positions for the particles. Note that the class label of an individual particle does not change by sampling.

%


\begin{algorithm}[th!]
 \caption{Tree Probability Rebuild}
\begin{algorithmic}[1]
\REQUIRE Tree structure $\mathcal{T}$, tree level $\mathcal{C}_b$, particle set $X_b$
\STATE  Update the probabilities of classes $c_i\in\mathcal{C}_b$ from $X_b$ weights,
 \[\mathbf{P}^t(c_i)=\frac{\sum_{c(x)=c_i} w^t(x)}{\sum_{x\in X_b} w^t(x)}\]\label{step_rebuild}
\FOR{children of $\mathcal{C}_b$ recursively \TO the bottom of $\mathcal{T}$}{\STATE Update the probability of the child $\underline{c}$ relative to the probability of its parent $\bar{c}=\rho(\underline{c})$: \[\mathbf{P}^t(\underline{c})=\mathbf{P}^{t-1}(\underline{c})\frac{\mathbf{P}^t(\bar{c})}{\mathbf{P}^{t-1}(\bar{c})}\]}  \label{lin_children}
\ENDFOR
\FOR{parents of  ${\mathcal{C}}_0$ recursively \TO the top of $\mathcal{T}$}
 \STATE Update the probability of the parent $\bar{c}$ relative to the probability of its children $\underline{c}=\rho^{-1}(\bar{c})$: \[\mathbf{P}^t(\bar{c})=\sum_{\underline{c}=\rho^{-1}(\bar{c})} \mathbf{P}^t(\underline{c})  \;\;\]  \label{lin_parents}
\ENDFOR
\end{algorithmic}
\label{alg:tree_rebuild}
\end{algorithm}


\subsection{Observations}
 A coarse observation $\xi^t\in\mathcal{C}$ of level $b_\xi$ relates to all the particles from classes that are \emph{alive} at that level of the hierarchy, ${\mathcal{C}}_\xi=\{c_i\in\mathcal{C}|\,\,b_i\leq b_\xi< d_i\}$. 

\begin{figure}[th!]
\centering
\includegraphics[width=0.6\textwidth]{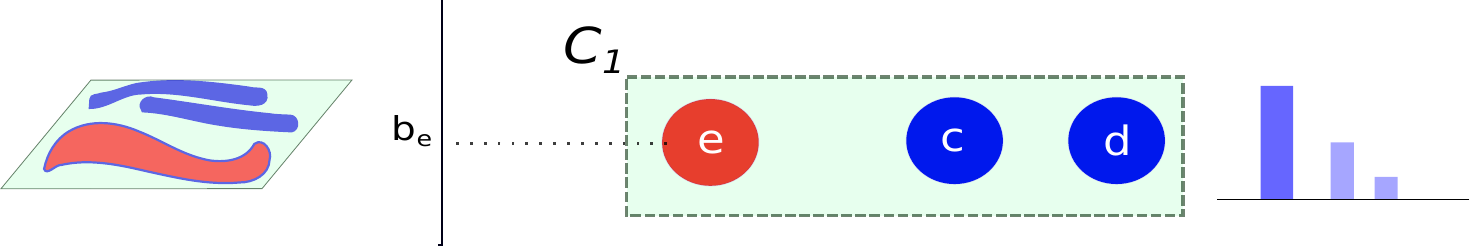}
\caption{A coarse observation targets a layer in the tree, and the particles of all the classes that are alive will be updated using tree class distance.} 
\end{figure}

To update a  particle $x(z^t,c)$ of a class $c\in {\mathcal{C}}_\xi$ we use the \emph{tree class distance} between $c$ and $\xi^t$, which we define for two classes $c_1$ and $c_2$ as 
 the birth index of the first shared parent of $c_1$ and $c_2$ in the tree. This distance measures how far we have to climb in the tree for the two classes to be similar enough and join the same cluster, or alternatively how large the $\epsilon$-balls around the points of one class need to be to include the other. For example, the  class distance between class $e$ and class $g$ in Figure~\ref{fig:overview}  is 
$b_g$, which is also the case for classes $c$ and $e$. The weight of a particle is then updated relative to the tree distance, $ w\propto -\log(\delta_{\mathcal{T}}(\xi^t,c))$.

On the other hand, updating a particle with a position observation is straightforward, relative to the Euclidean distance between the observation and the particle's position, $ w\propto -\log(\delta_E(\xi^t,z))$.


\begin{algorithm}[th!]
 \caption{Observation Update}
\begin{algorithmic}[1]
\REQUIRE  observation $\xi^t$, tree structure $\mathcal{T}$
\IF{fine observation}{
\STATE $\hat{\mathcal{C}}=\mathcal{C}_0$, $\hat{X}=X_0$
\FOR{every particle $x(z,c) \in\hat{X}$}
{
\STATE update weight relative to Euclidean distance to $\xi^t$: $w^t\propto -\log(\delta_E(z,\xi^t)).$
}
\ENDFOR
}
\ELSIF{coarse observation}
{
\STATE Identify the level $b_\xi$ of the observation  $\xi^t$
\STATE Find all the classes that are alive at the level: $\hat{\mathcal{C}}=\{c_i\in\mathcal{C}:\,\,b_i\leq b_\xi< d_i\}.$
\FOR{every class $c \in\hat{\mathcal{C}}$}
{
\STATE find the tree distance $\delta_{\mathcal{T}}(c,\xi^t)
,$ the birth index of the first shared parent of $c$ and $\xi^t$.
\FOR{every particle $x(z,c) \in X_c$}
{
\STATE update weight relative to tree distance to $\xi^t$: $w^t\propto -\log(\delta_{\mathcal{T}}(c,\xi^t)).$
}
\ENDFOR
}
\ENDFOR
}\ENDIF
\STATE Rebuild the tree probabilities from $\hat{\mathcal{C}}$ (Algorithm~\ref{alg:tree_rebuild}).
\FOR{every particle $x(z,c) \in X_{\mathcal{C}\setminus\hat{\mathcal{C}}}$}
\STATE Update particle weights: $w^t=w^{t-1}\frac{\mathbf{P}^t(c)}{\mathbf{P}^{t-1}(c)}$ \label{lin_rest_particles}
\ENDFOR
\end{algorithmic}
\label{alg:obs_update}
\end{algorithm}


 Then, the probabilities of the classes of $\mathcal{C}_\xi$ are recomputed as the  sum of their  particles' normalised weights. Note that the coarse update is qualitative in nature such that all the particles of a certain class from $\mathcal{C}_\xi$would get the same update regardless of the particle positions.

The updated class probabilities of ${\mathcal{C}}_\xi$ propagate to the rest of the tree as in Algorithm~\ref{alg:tree_rebuild}. At this stage, children of updated classes are updated first recursively relative to their parents' new probabilities, \[\mathbf{P}^t(\underline{c})=\mathbf{P}^{t-1}(\underline{c})\frac{\mathbf{P}^t(\bar{c})}{\mathbf{P}^{t-1}(\bar{c})},\; \forall \underline{c}=\rho^{-1}(\bar{c}).\] 

Then, the updates propagate upwards by updating all parents recursively, summing up their children's probabilities, \[\mathbf{P}^t(\bar{c})=\sum_{\underline{c}=\rho^{-1}(\bar{c})} \mathbf{P}^t(\underline{c})  \;\;\]

 Once the tree probabilities are balanced, the rest of the particle weights are updated to reflect their new class probabilities, $w^t=w^{t-1}\frac{\mathbf{P}^t(c)}{\mathbf{P}^{t-1}(c)},\;\forall x(.,c): c\in\mathcal{C}\setminus\mathcal{C}_\xi$.

%
%
%
%

The final step is to resample $N$ particles from the finest level of the tree $X_0$ with equal weights to get the posterior particle set after incorporating the evidence. To guard against particle depletion, we randomly replace the classes of $v$ of the particles uniformly randomly to classes from $\mathcal{C}_0$. From this new particle set the process repeats.


\section{Experiments}
We evaluate the performance of {\bf MHPF} in a number of 2-dimensional navigation domains over two baselines which are particle filters without access to the hierarchical structure.  {\bf BL1} is a basic particle filter~\cite{probabilistic_robotics} with $N$ particles $x(z,c)$, with $c$ restricted to the single trajectory class $\mathcal{C}_0$. Thus, each particle follows its single-trajectory class. Secondly, {\bf BL2} is a particle filter with $N$ particles $x(z)$ where all the particles follow the localised dynamics of the combination of all the trajectories with $\kappa$ noise. Note that {\bf BL1} is equivalent to the bottom layer in the MHPF filter stack, and {\bf BL2} is equivalent to the top layer.

We show the improvements using a number of metrics. We use the mean squared error of the filter's point prediction to show efficacy, and we evaluate performance by showing the distance of the filter's predicted class to the ground truth as well as the time needed to converge to the true class being followed by the agent.

We use synthetic datasets as well as real world data in the experiments. Each experiment runs over 10 randomly selected scenarios described by corresponding ground truth trajectories that the process follows. The trajectories are uniformly discretised, and the length of a trial will depend on the number of points in the discretisation. Each trial is repeated 25 times and the results are averaged. 

 Each of the experiments had $N=100$ particles at any of the tree levels. At every time step, observations are generated from the discretised ground truth. A fine observations is defined as $z^t+(\gamma_1,\gamma_2)$ where $z^t\in\mathds{R}^2$ is ground truth at time $t$, $\gamma_1,\gamma_2\in\mathds{R}$ are chosen uniformly randomly from $[0,\psi]$ where $\psi$ is the observation noise parameter of the experiment.  A coarse observation is generated by sampling a number of points ($n=10$) from $\mathcal{N}(z^t,\psi^2)$ where $\mathcal{N}$ is a normal distribution, $z^t$ is the ground truth at time $t$ and $\psi$ is the observation noise parameter. Then, the class that has the highest probability to generate these samples is chosen as $\xi^t$. For dynamics we used a localised model as in Section~\ref{localised_dynamics} with $\epsilon$ (the size of the $\epsilon$-ball) equals to $b_c$ for some coarse class $c$ and with a noise term from $[0,\kappa]$ for the dynamics noise parameter $\kappa$. We used KD-trees for efficient selection of neighbourhood points. At the end of every step, $v=1\%$ of the particles is changed randomly.

\subsection{Synthetic datasets}
\begin{figure}[th!]
\centering
\begin{subfigure}{0.3\textwidth}
\fcolorbox{lightgray}{white}{\includegraphics[width=0.8\textwidth]{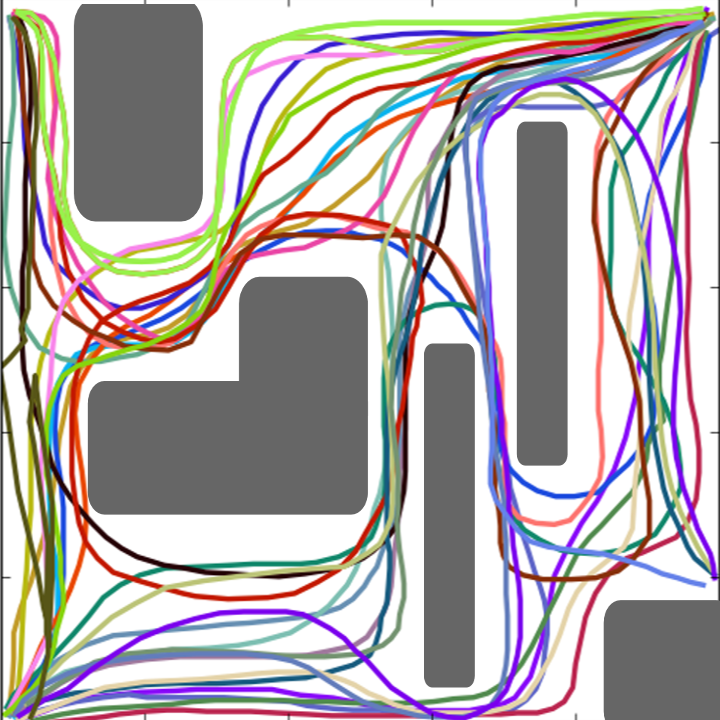}} 
\end{subfigure} \hspace{5mm} \begin{subfigure}{0.3\textwidth}
\fcolorbox{lightgray}{white}{\includegraphics[width=0.8\textwidth]{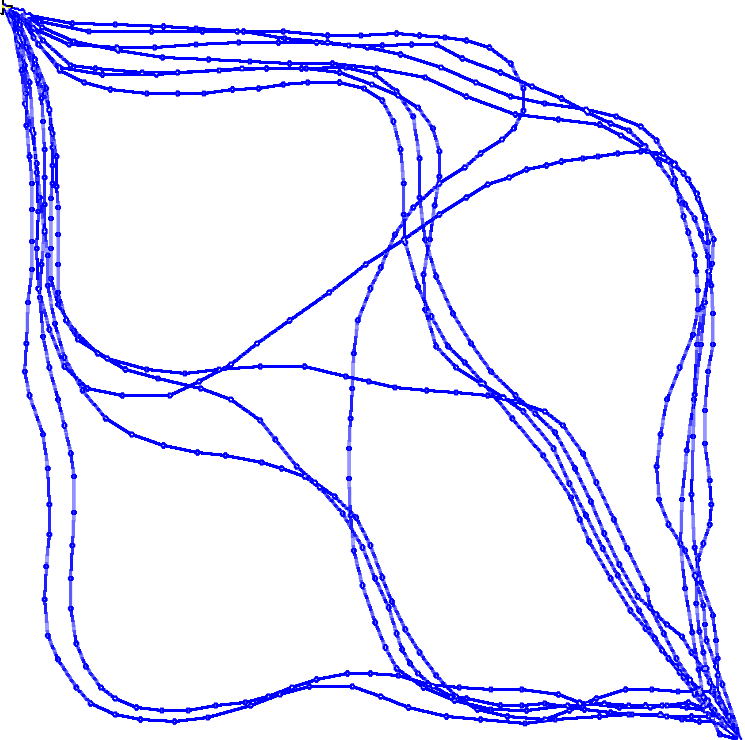}} 
\end{subfigure}
\caption{\footnotesize Synthetic datasets of  33 trajectories in a 2D configuration space with obstacles, and 13 trajectories that starts and ends at the same position in a 2D plane.\label{fig:blobs}}
\end{figure}

We work with two synthetic domains, the first represents a 2-dimensional configuration space with 33 trajectories with general start and end positions, and the second one has 13 trajectories with a fixed start and end position (Figure~\ref{fig:blobs}).

 For the configuration space dataset, we compute the filter's predicted position at time $t$ as the $w$-weighted average of the particle positions, and report  the average of the mean squared error (MSE) of the ground truth to this predicted position over time and for 10 random scenarios, each repeated 25 times. {\bf MHPF} achieved a mean of 0.27 (standard deviation of 0.04) compared to {\bf BL1} which achieved 0.38(0.14) and {\bf BL2} which achieved 0.53(0.13).   This experiment uses fine observations only.

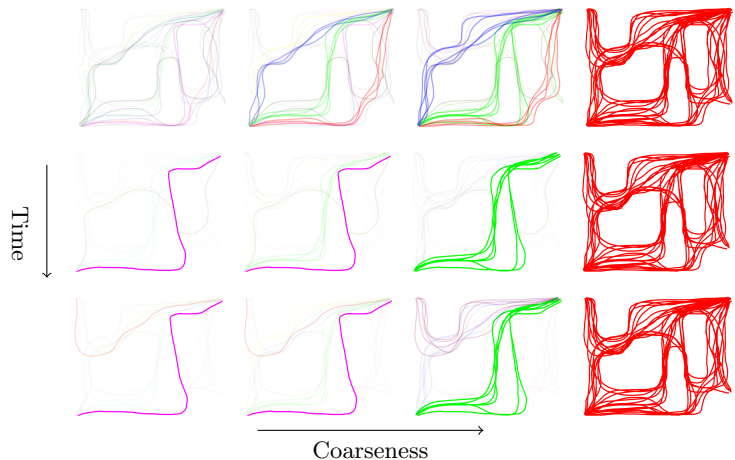
\begin{figure}[th!]
\centering
\begin{subfigure}{0.8\textwidth}  
  \centering
    \hspace{0.08\textwidth}
    \includegraphics[width=0.22\textwidth,height=0.178\textwidth]{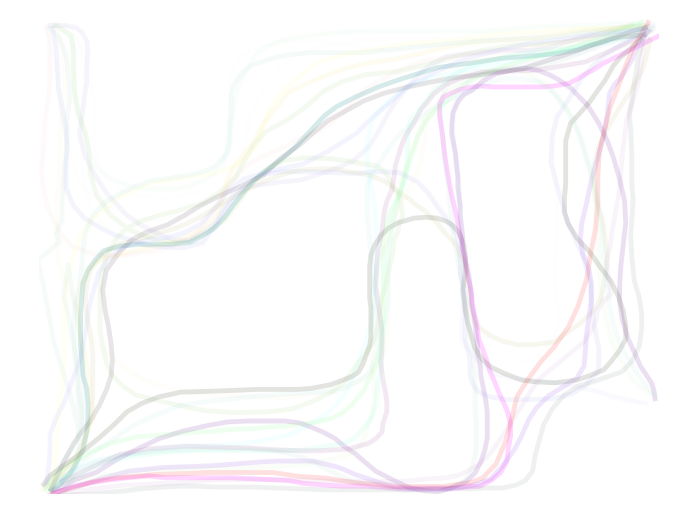}  \includegraphics[width=0.220\textwidth,height=0.178\textwidth]{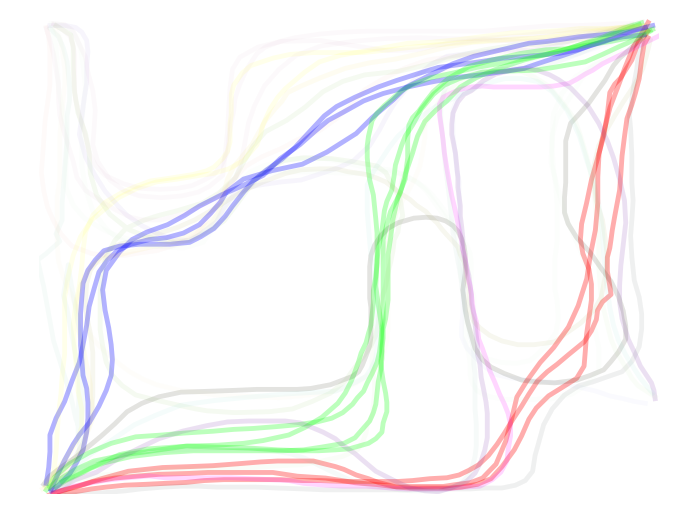}  \includegraphics[width=0.220\textwidth,height=0.178\textwidth]{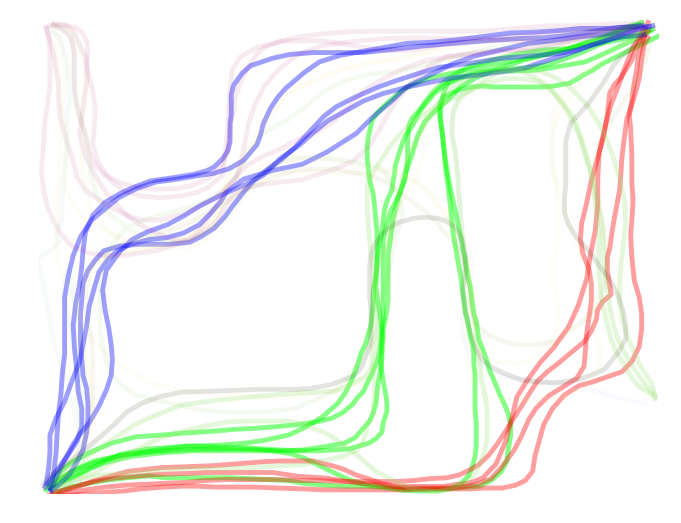}  \includegraphics[width=0.220\textwidth,height=0.178\textwidth]{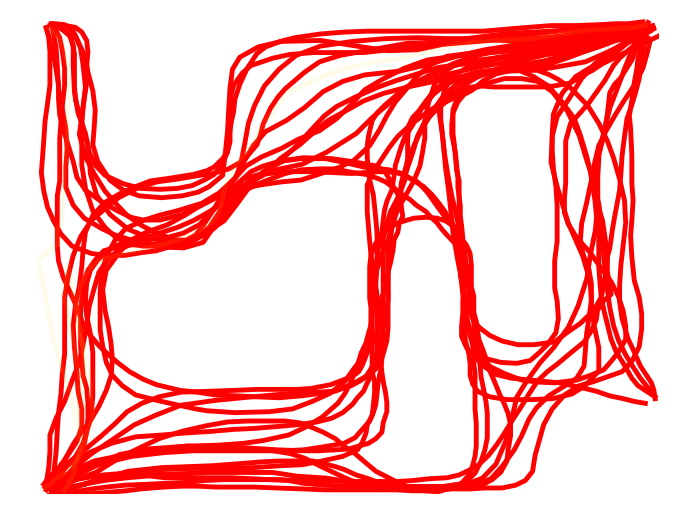} 
\end{subfigure}%

\vspace{0.15cm}
\begin{subfigure}{0.8\textwidth}  
  \centering
   
    \begin{tikzpicture}\node[label={[label distance=0.0cm,text depth=-0.2ex,rotate=-90]left:Time}] at (1.5,0) {};\end{tikzpicture} \begin{tikzpicture}[scale=1.5]\draw[->] (0,1) -- (0,0);\end{tikzpicture}\hspace{0.012\textwidth}
    \includegraphics[width=0.220\textwidth,height=0.178\textwidth]{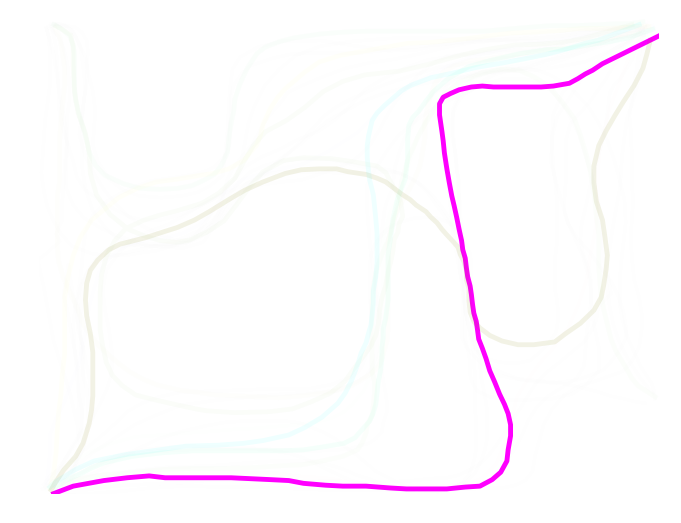}  \includegraphics[width=0.220\textwidth,height=0.178\textwidth]{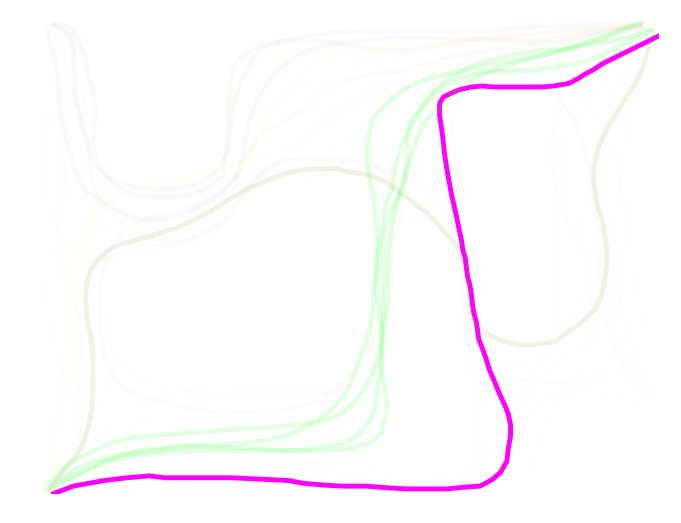}  \includegraphics[width=0.220\textwidth,height=0.178\textwidth]{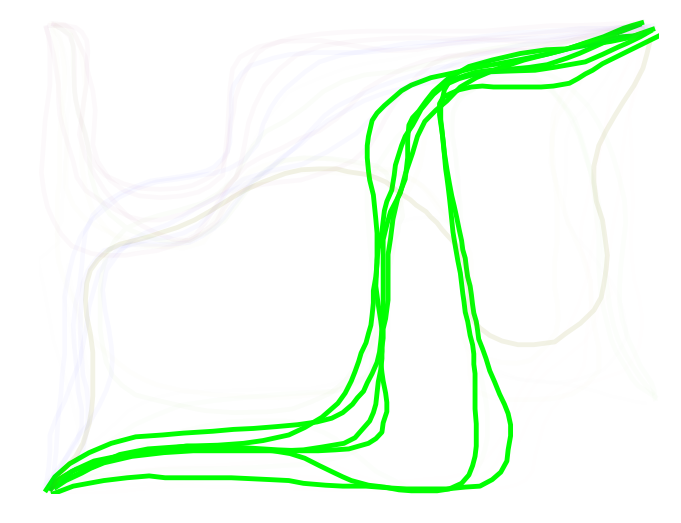}   \includegraphics[width=0.220\textwidth,height=0.178\textwidth]{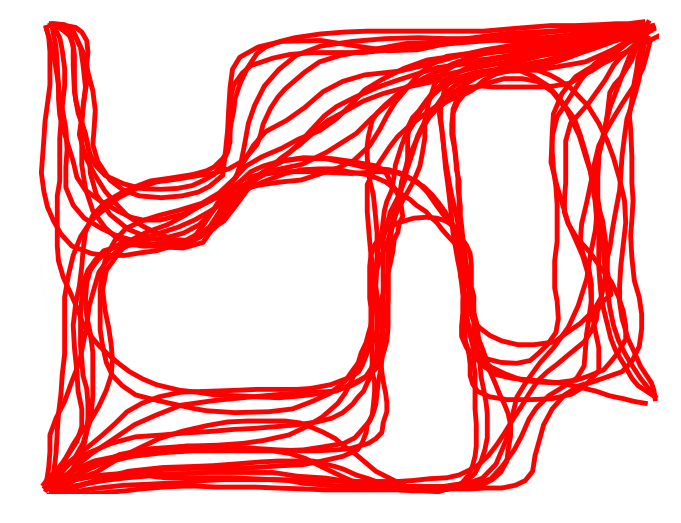} 
\end{subfigure}%

\vspace{0.15cm}
\begin{subfigure}{0.8\textwidth}  
  \centering 
    \hspace{0.072\textwidth}
    \includegraphics[width=0.220\textwidth,height=0.178\textwidth]{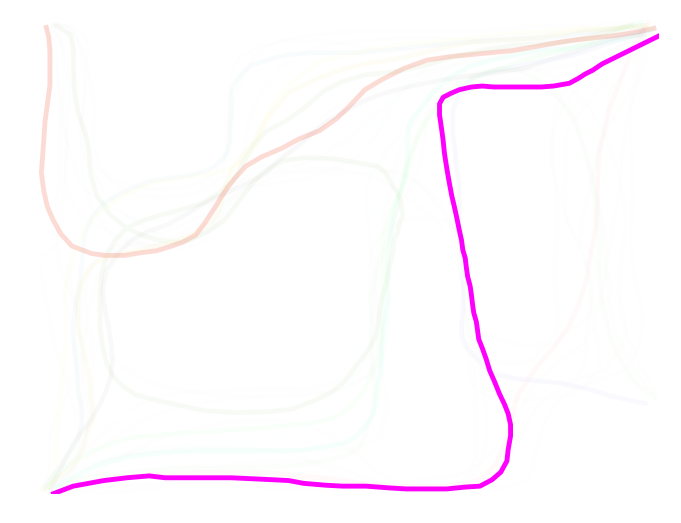}  \includegraphics[width=0.220\textwidth,height=0.178\textwidth]{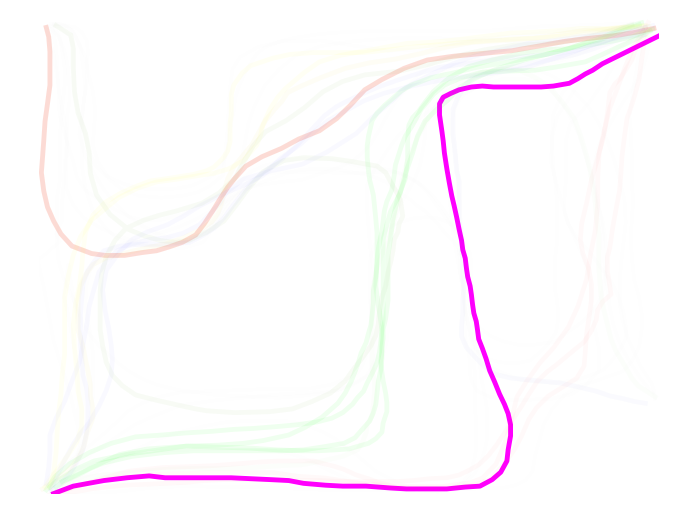}  \includegraphics[width=0.220\textwidth,height=0.178\textwidth]{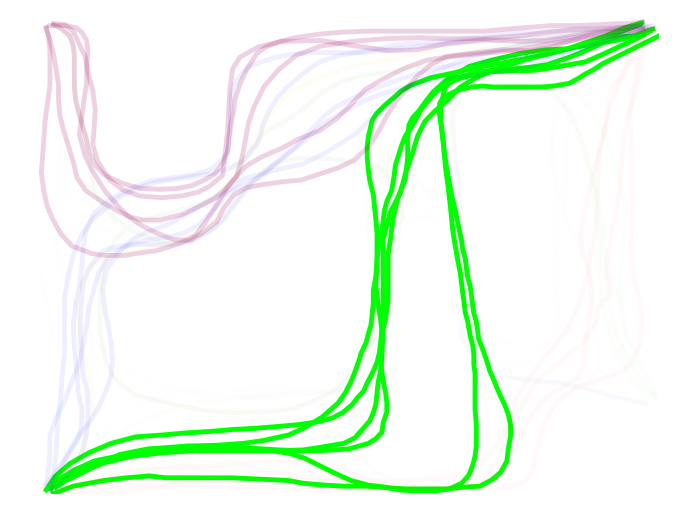}   \includegraphics[width=0.220\textwidth,height=0.178\textwidth]{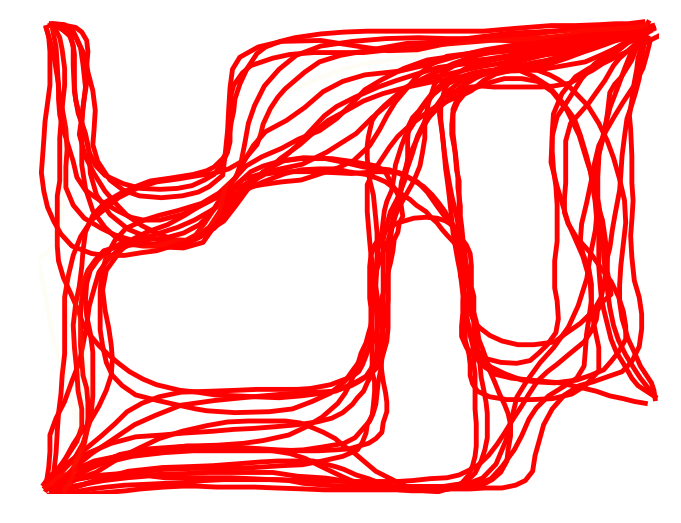} 
\end{subfigure}%

\vspace{0.06cm}
 \begin{tikzpicture}[scale=3]\draw[->] (0,0) -- (1,0);\end{tikzpicture}\\
\begin{tikzpicture}\node[label=Coarseness] at (2,1.5) {};\end{tikzpicture}

\caption{\footnotesize Evolution of the filter's prediction. The columns are for different levels across the tree with the finest at the left. The rows are for different time steps with the first at the top. Each panel shows the trajectories of the alive classes. The opacity of the line reflects the probability of the class. }
\label{fig:topn_mouse33}
\end{figure}

Figure~\ref{fig:topn_mouse33} illustrates the kind of multi-resolution output the filter can produce. It shows the evolution of the filter's \emph{maximum a posteriori} (MAP) class with time and for different levels of the tree. Each column shows the classes of some level $b$, with the leftmost column showing the finest level with individual trajectory classes and the rightmost column showing the coarsest level (a single class combining all the trajectories), while rows show progress over time. The thicker the trajectories are, the more likely their class is.

\begin{figure}[th!]
\centering
\begin{subfigure}{0.5\textwidth}\vspace{-3mm}
\includegraphics[width=\textwidth]{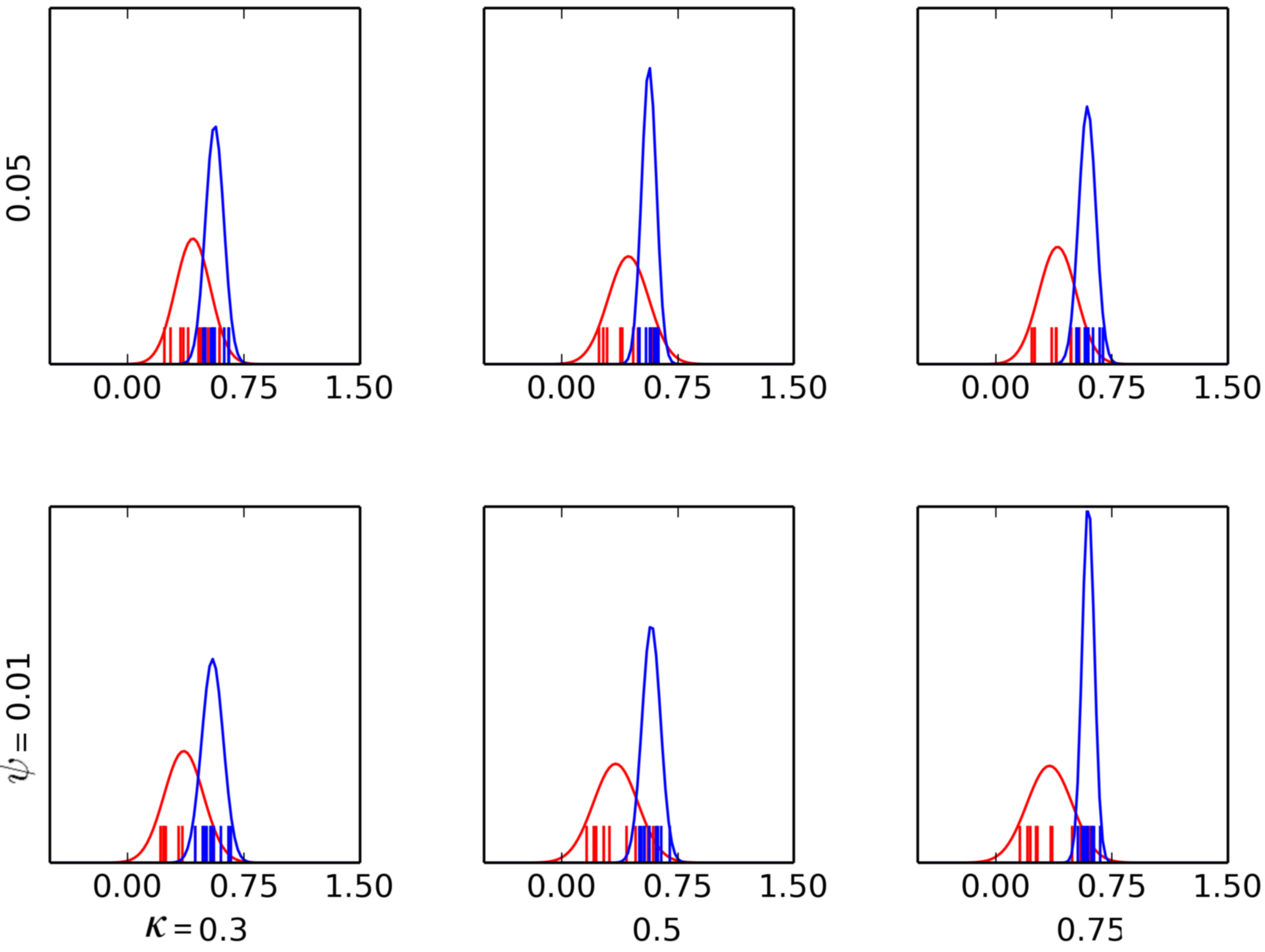} 
\caption{Distance between the ground truth and MAP prediction of MHPF~(red) and BL1~(blue). Coarse instructions were provided stochastically 50\% of the time. The plot shows the robustness against noise as dynamics noise ranges between 30\% to 75\%, and observation noise ranges between 1\% to 5\%. MHPF converges to a better solution compared to the baseline, and this result is statistically significant at a p-value of 0.004. \label{fig:mpf_dist}} 
\end{subfigure}\hspace{3mm}\begin{subfigure}{0.45\textwidth}
\includegraphics[width=\textwidth]{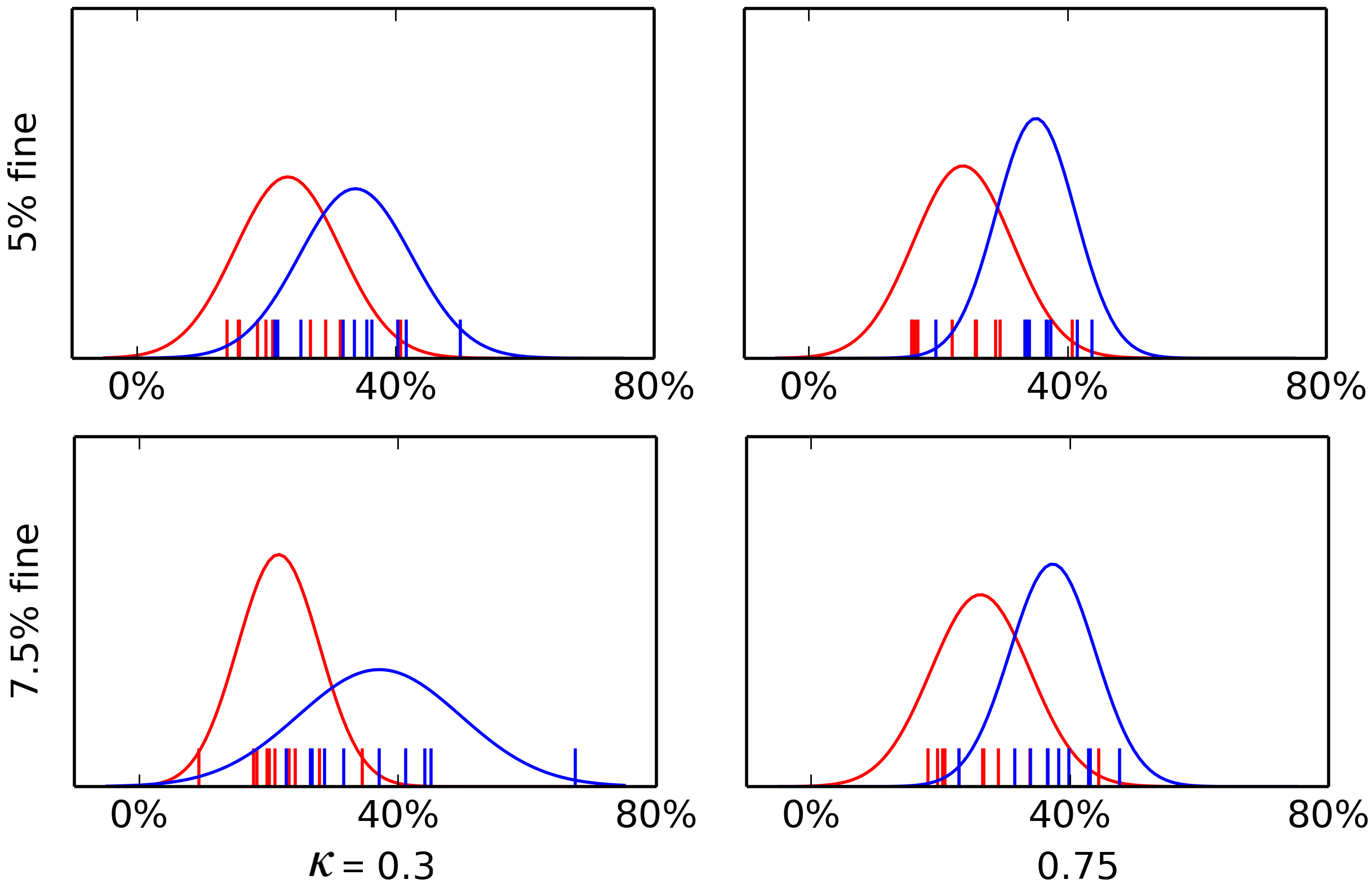} 
\caption{Time needed to reach within 33\% of convergence for MHPF~(red) and BL1~(blue). Fine observations are provided for a lead-in period (5\% and 7.5\% of trial time) then only coarse observations are given. The plot shows the benefit of coarse observations to convergence time. Dynamics noise ranges from 30\% to 75\%, and observation noise is set to 1\%. MHPF converges faster to the correct solution than the baseline, and this result is statistically significant at a p-value of 0.02. \label{fig:mpf_time}} 

\end{subfigure}

\caption{Performance results in the synthetic domains comparing MHPF~(red) and BL1~(blue). X axis shows dynamics noise. The experiments  were run on 10 different test scenarios, each repeated 25 times, then the results were averaged.}

\end{figure}

Next, using the 13 trajectory dataset, we compare {\bf MHPF} with {\bf BL1} in a situation where, in addition to the consistent fine observations, coarse observations are produced stochastically 50\% of the time. This is motivated by use cases where high-level qualitative information (e.g. human instructions) might exist along the finer localised measurements. We analyse the benefit of this additional knowledge by plotting the average tree class distance of the MAP prediction of the filters to the ground truth. We show the results for different values of dynamics noise ($\kappa=30\%,50\%, 75\%)$ and different values of observation noise ($\psi=1\%, 5\%$).  The results are reported in Figure~\ref{fig:mpf_dist}.

Finally, using the same dataset, we analyse the situation where fine observations are only provided for a fraction of the time (a lead-in period of $5\%$ and $7.5\%$ of the trial length), then only coarse observations are given. We present the effect of that on the time needed by {\bf MHPF} and {\bf BL1} to converge by plotting the time needed for the class distance to reach within the 33\%-ball of the ground truth. We show the results for observation noise $\psi=1\%$ and for different values of dynamics noise ($\kappa=30\%, 75\%)$. The results are reported in Figure~\ref{fig:mpf_time}.

%

\subsection{Tanker vessel data}
\begin{figure}[th!]
\centering

\begin{subfigure}{0.40\textwidth}
\includegraphics[width=\textwidth]{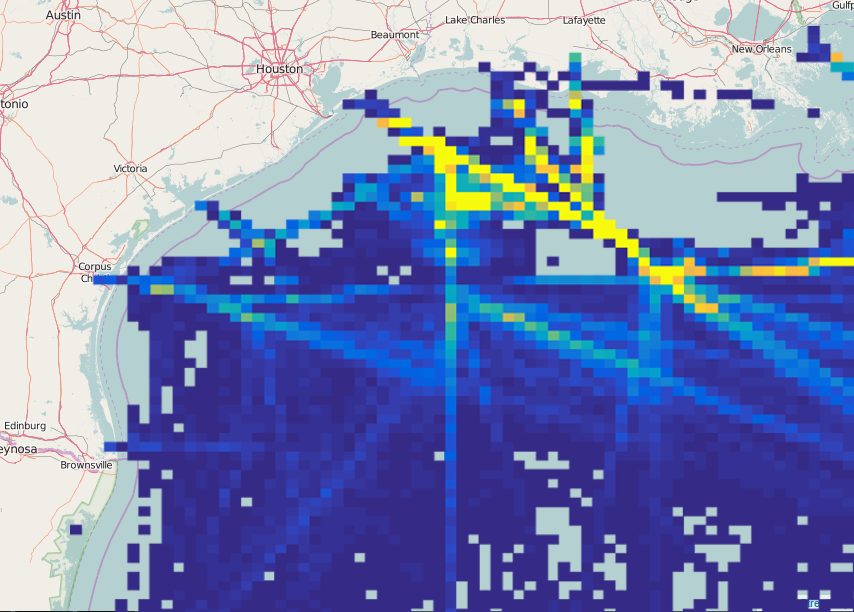}  
\end{subfigure} \hspace{0.5cm} \begin{subfigure}{0.40\textwidth}
\fcolorbox{lightgray}{white}{\includegraphics[width=0.5\textwidth]{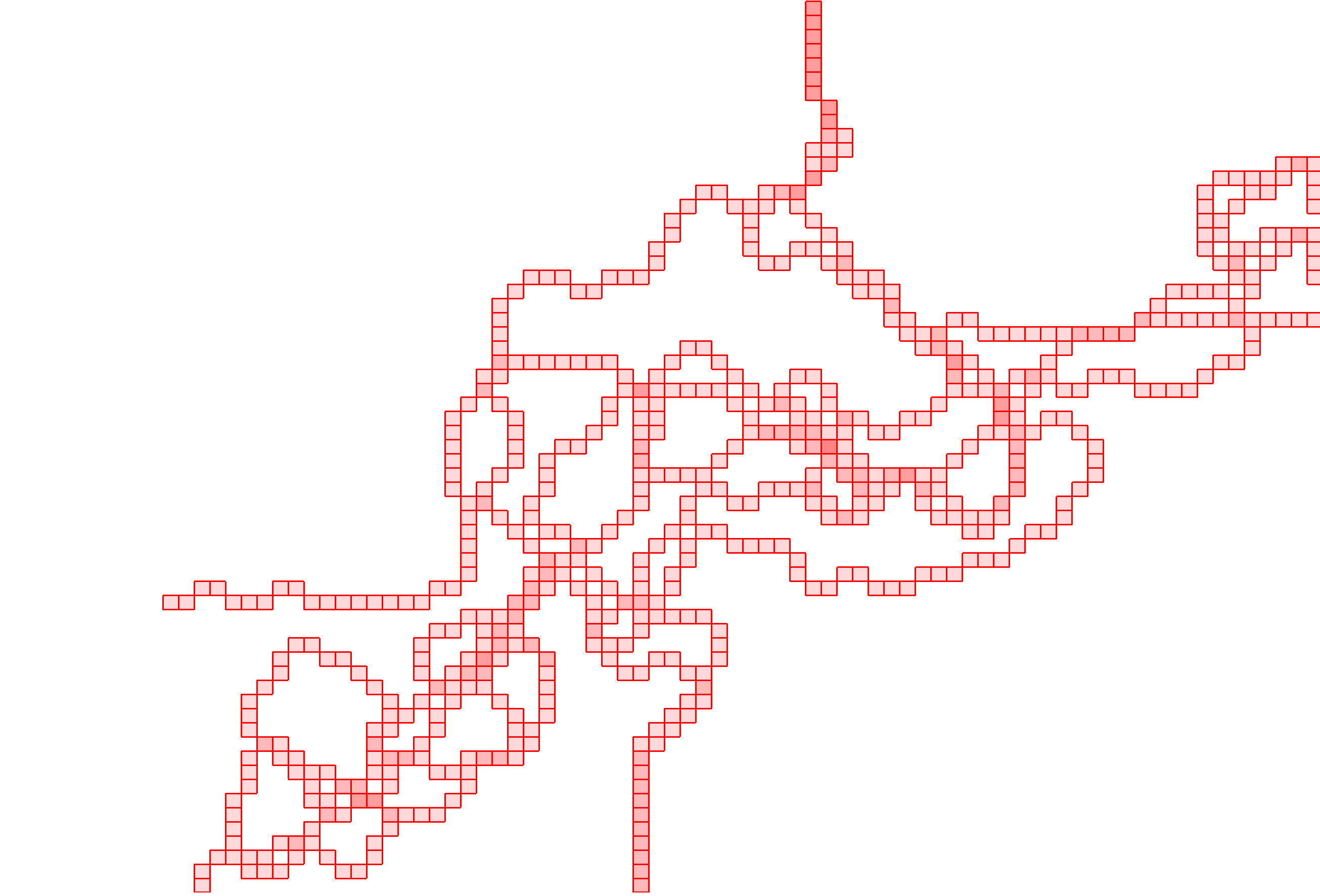}}\fcolorbox{lightgray}{white}{\includegraphics[width=0.5\textwidth]{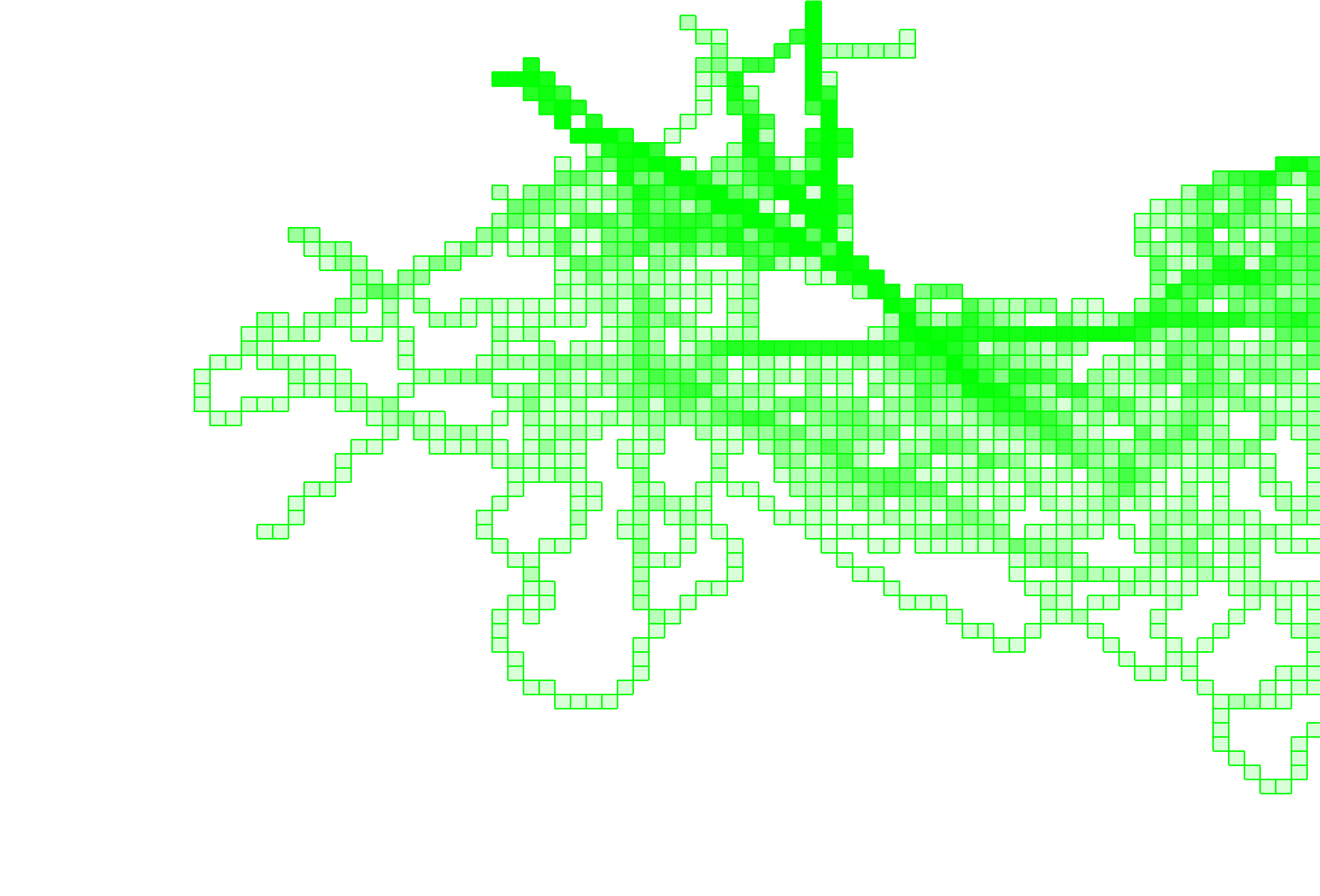}}\\
\fcolorbox{lightgray}{white}{\includegraphics[width=0.5\textwidth]{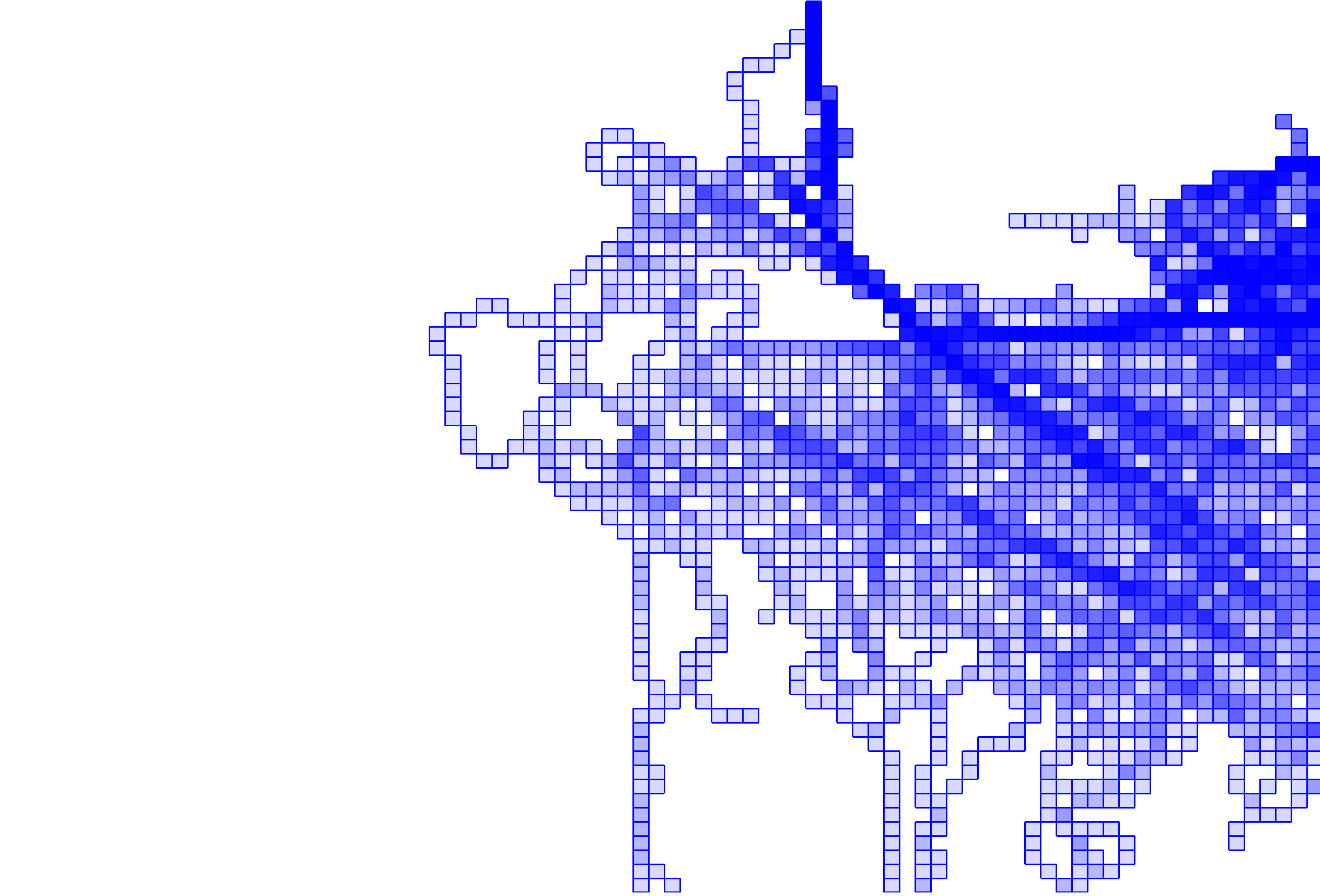}}
\end{subfigure}%

\caption{\footnotesize Density and the trajectory classes at one level in the tree in the tanker vessel experiment\label{fig:tanker}}
\end{figure}

This experiment uses publicly-available data regarding the movement of ships in a harbour area. Specifically, we utilise records of tanker vessel tracks around the Gulf of Mexico~\cite{marine_cadastre}. From the data which is available in the form of density/occupancy graph we generate 194 trajectories by weighted random walks from manually-selected initial positions, such that a trajectory is more likely to follow the denser areas and does not change direction much often. Figure~\ref{fig:tanker} shows the density and the trajectory classes for $b=70$.

\begin{figure}[th!]
\centering
\includegraphics[width=0.7\textwidth]{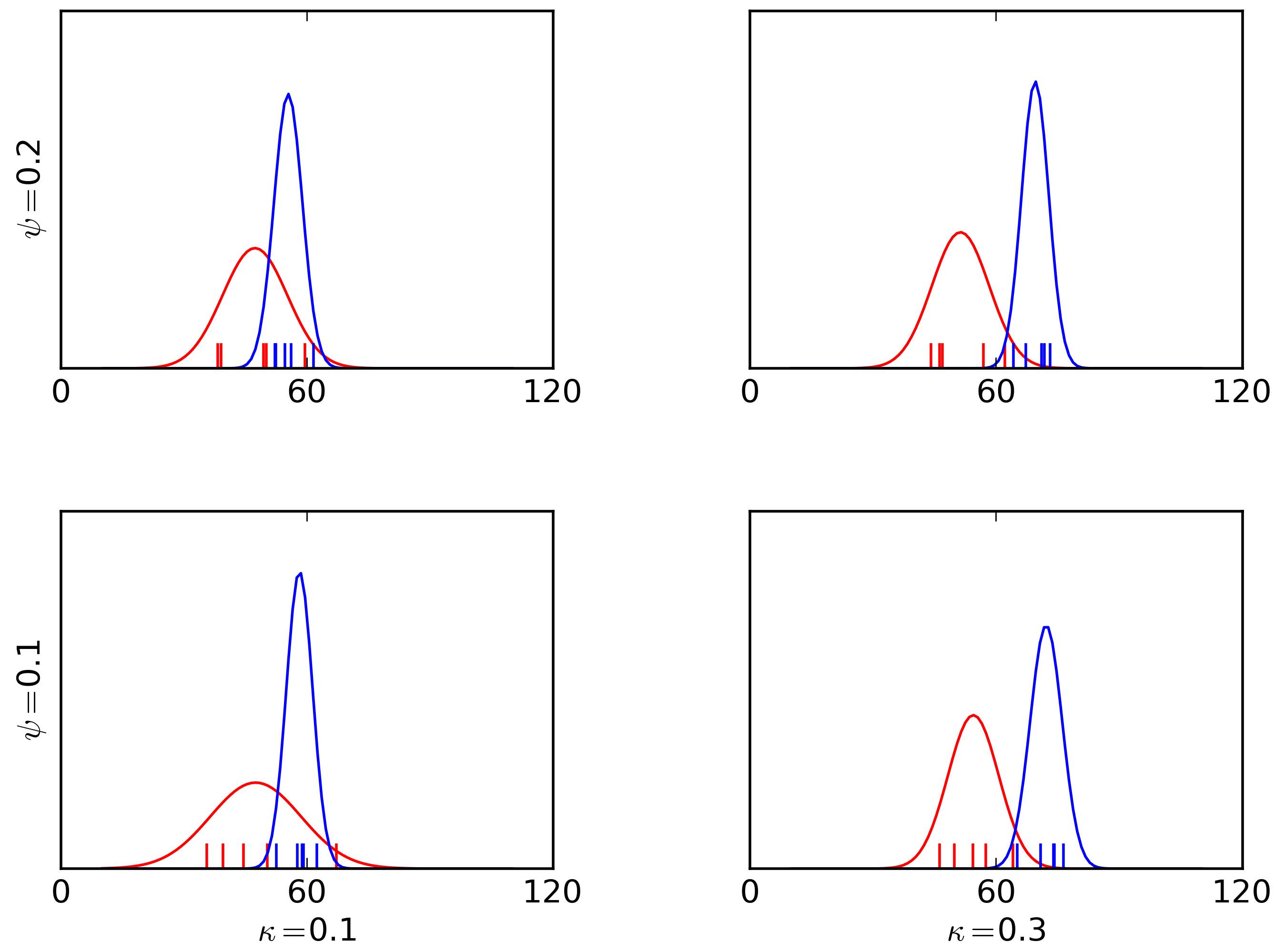} 
\caption{Distance between the ground truth and the MAP prediction of MHPF (red) and the basic filter (blue). Coarse instructions were provided stochastically 50\% of the time. Dynamics noise ranges from 10\% to 30\%, and observation noise from 10\% to 20\%. The experiment was run on 5 different test scenarios, each repeated 25 times, then the results were averaged. The result is statistically significant with p-value of 0.1. \label{fig:tanker_mpf_dist}} 
\end{figure}

Compared to {\bf BL1}, we explore the benefit to convergence when receiving coarse information 50\% of the time along with fine observations in ship tracking scenarios. The reported values in Figure~\ref{fig:tanker_mpf_dist} are averages of tree class distance between the MAP prediction of the filter and the ground truth trajectory with dynamics noise ranging from 10\% to 30\% and observation noise ranging from 10\% to 20\%.

%
%
%

\section{Conclusion}
We propose a novel approach to utilising a hierarchical clustering over trajectories (a filtration) to devise a correspondingly hierarchical representation of probability distributions over the underlying state space so as to enable Bayesian filtering. A key benefit of our methodology is the ability to incorporate coarse observations in the estimation process to seamlessly allow for potential inhomogeneity in sensor readings, such as when a GPS device obtains position fixes with varying confidence, or for signals at varying degrees of coarseness, such as when a human user instructs a robot in relational terms. We demonstrate the usefulness of this technique with experimental domains of increasing complexity, ranging from a synthetic data set intended to illustrate the elements of the operation of this algorithm to real data drawn from tracked vessels in a harbour environment. We show that the proposed algorithm is able to perform much better than a more conventional particle filtering procedure through the use of the hierarchy, and also that it is able to make use of observations that are presented in a form that would be hard to reconcile with the way conventional particle filtering schemes are constructed. We view this work as a step towards systems with more flexible predictive modelling ability in interactive settings, something that is becoming increasingly more prevalent as robots cohabit human-centred environments.

\bibliographystyle{splncs}
\bibliography{ref}

\end{document}

%% file: Ram_introduction.tex
Autonomous agents that act in dynamic environments, including humans and other active agents, must possess the capacity to make predictions about the changes in the environment and correspondingly take actions with respect to their best estimate about the state of the world. In particular, there is an increasing need for robots to make predictions about the activity of other agents - what is the state of the activity of the other agent and what might they do next? 

Traditionally, these prediction problems have been addressed using tools from \emph{state estimation theory}, which is most mature when the source of uncertainty is rooted in the dynamics and noise of sensorimotor processes. Particle filters are used extensively in robotics when there is the need to perform Bayesian state estimation in nonlinear systems with noisy and partial information about the underlying state~\cite{isard1998ijcv,Doucet2001,probabilistic_robotics}. In this approach, the posterior probability distribution over the states given a sequence of measurements is approximated with a set of particles which represent state hypotheses.

When the hypotheses pertain to spatial activity, such as navigation of people/robots in typical human-centred environments, or perhaps in more complex configuration spaces such as when dealing with full body motion, the underlying dynamics are best described in a hierarchical fashion, so that a movement is not just determined by the local laws of dynamics and noise characteristics, but also by longer term goals and preferences. This has at least two important implications for predictive models. We need techniques such as for activity estimation to be able to accept evidence at varying scales - ranging from very precise position measurements to coarser forms of human feedback (e.g., ``she is heading to the right between two obstacles'') ~\cite{linguistic,kollar2014} or variable resolution sensory signals (e.g., GPS-like devices that provide location estimates to within a spatial neighbourhood of variable extent). Correspondingly, we would like to be able to output predictions at multiple scales, to support decision making at all these levels. These 
desiderata form the primary focus of this paper, in which we propose a technique that accepts evidence and provides estimates at multiple resolutions.

There is a long tradition of hierarchical modelling of motion which could inform the design of such techniques. Early models of large-scale spatial navigation~\cite{ssh} considered ways in which multiple representations, ranging from coarse and intuitive topological notions of connectivity between landmarks to a more detailed metrical and control level description of action selection, could be brought together in a coherent framework and implemented on robots. Other recent methods, driven more from motion planning and control considerations e.g.~\cite{Belta2007,Burridge01061999}, propose ways in which control vector fields could be abstracted so as to support reasoning about the hybrid system that is aimed at solving the larger-scale tasks. 

While these works provide useful inspiration, they do not directly address our aforementioned desiderata. Firstly, the hierarchy is often statically defined by the designer of the system and the algorithm. In many applications, it is of interest to be able to learn these from data - both because this enables online and continual adaptation over time, but also because the underlying principles determining the types of motion may not be fully understood (e.g., the activity of people in a complex environment, driven by private and varied utility functions). Secondly, the approaches that are principled in the way they define the hierarchy are often silent on how best to integrate with the methodology for maintaining Bayesian belief estimates, such as with a particle filter - the question of how best to define a correspondingly-hierarchical activity estimation method is largely open.

There is indeed prior work on the notion of hierarchy in state estimation with particle filters. For instance, Verma et al.~\cite{ram2} define a variable resolution particle filter for operation in large state spaces, where chosen states are lumped into aggregated states so that the complexity of the particle filter may be reduced. Brandao et al.~\cite{ram1} devise a subspace hierarchical particle filter wherein the focus is on defining subspaces for which state estimation calculations can be run in parallel, alleviating the computational burden through factored parallel computation. There are a variety of ways in which computations have been factored~\cite{similarity}, e.g., by partitioning the ways in which sampling calculations may be performed for tracking articulated objects with an implied hierarchy~\cite{ram3}, of by using a hierarchy of feature encodings~\cite{cascade} and contour detection~\cite{contour}.

In this paper, we take a different approach to learning the hierarchy, using which we devise a novel construction of a bank of particle filters - one at each scale - maintaining consistent beliefs over the trajectories as a whole (in the spirit of plan and activity recognition) and through that over the state space. Given a set of trajectories (such as from historical observations of activity) and a notion of trajectory similarity, we define procedures for hierarchical clustering of these trajectories. The output of clustering is a tree-structured representation of trajectory classes that correspond to incrementally-coarser partitions of the underlying space. We present an agglomerative clustering scheme~\cite{hierarchical_clust} using the Fr\'{e}chet distance between trajectories~\cite{discrete_frechet}. This construction of trajectory clustering in the form of a filtration is inspired by earlier work using persistent homology~\cite{carlsson2009topology,RSS_14,IJRR}, but is instantiated in a simpler and computationally more efficient manner through the 
use of Fr\'{e}chet distance based agglomeration. Equipped with this data-driven notion of a hierarchy, we show how to define the dynamics at the different levels, and how they can be employed with a new stream of observations to provide probability updates over time and over the classes in the tree.

Our construction of the filter allows us to fluently incorporate readings of varying resolution if they were accompanied by an indication of the coarseness with which the observation is to be interpreted. This issue of variability is much broader in scope, covering many other aspects of dynamical systems behaviour~\cite{nudging,weather2}. 

We evaluate our proposed method by showing how unsupervised learning of hierarchical structure in the activity data enables the bank of particle filters at multiple scales (which we refer to more concisely, with slight abuse of terminology, as a `multiscale particle filter') to perform better than  baselines both in terms of normalised error in predicting the position of an agent with respect to the ground truth trajectory, and in terms of the time taken for the belief to converge to the true trajectory of class (depending on the resolution of the prediction being considered). We perform such experiments first with a synthetic dataset which brings out the qualitative behaviour of the procedure in a visually intuitive manner, and then with a real world dataset based on tracks of ships in a harbour (based on a database associated with the worldwide AIS system).

%% file: tree3_1.pdf_tex
\begingroup%
  \makeatletter%
  \providecommand\color[2][]{%
    \errmessage{(Inkscape) Color is used for the text in Inkscape, but the package 'color.sty' is not loaded}%
    \renewcommand\color[2][]{}%
  }%
  \providecommand\transparent[1]{%
    \errmessage{(Inkscape) Transparency is used (non-zero) for the text in Inkscape, but the package 'transparent.sty' is not loaded}%
    \renewcommand\transparent[1]{}%
  }%
  \providecommand\rotatebox[2]{#2}%
  \ifx\svgwidth\undefined%
    \setlength{\unitlength}{785.77456841bp}%
    \ifx\svgscale\undefined%
      \relax%
    \else%
      \setlength{\unitlength}{\unitlength * \real{\svgscale}}%
    \fi%
  \else%
    \setlength{\unitlength}{\svgwidth}%
  \fi%
  \global\let\svgwidth\undefined%
  \global\let\svgscale\undefined%
  \makeatother%
{\tiny
  \begin{picture}(1,0.44499233)%
    \put(0,0){\includegraphics[width=\unitlength]{tree3_1.pdf}}%
    \put(0.70125671,0.22711074){\color[rgb]{0,0,0}\makebox(0,0)[lb]{\smash{\emph{$\mathcal{C}_0$}}}}%
    \put(0.46150411,0.2925487){\color[rgb]{0,0,0}\makebox(0,0)[lb]{\smash{\emph{$\mathcal{C}_e$}}}}%
    \put(0.65182007,0.3350083){\color[rgb]{0,0,0}\makebox(0,0)[lb]{\smash{\emph{$\mathcal{C}_f$}}}}%
    \put(0.59528754,0.3882223){\color[rgb]{0,0,0}\makebox(0,0)[lb]{\smash{\emph{$\mathcal{C}_g$}}}}%
    \put(0.48062282,0.18892431){\color[rgb]{1,1,1}\makebox(0,0)[lb]{\smash{\bf{a}}}}%
    \put(0.53789896,0.18944816){\color[rgb]{1,1,1}\makebox(0,0)[lb]{\smash{\bf{b}}}}%
    \put(0.59968162,0.18892431){\color[rgb]{1,1,1}\makebox(0,0)[lb]{\smash{\bf{c}}}}%
    \put(0.65846414,0.18944816){\color[rgb]{1,1,1}\makebox(0,0)[lb]{\smash{\bf{d}}}}%
    \put(0.51057804,0.25947224){\color[rgb]{1,1,1}\makebox(0,0)[lb]{\smash{\bf{e}}}}%
    \put(0.62650676,0.30425892){\color[rgb]{1,1,1}\makebox(0,0)[lb]{\smash{\bf{f}}}}%
    \put(0.56955423,0.36228692){\color[rgb]{1,1,1}\makebox(0,0)[lb]{\smash{\bf{g}}}}%
    \put(0.42592867,0.43420797){\color[rgb]{0,0,0}\makebox(0,0)[lb]{\smash{\emph{b}}}}%
    \put(0.3906589,0.26444362){\color[rgb]{0,0,0}\makebox(0,0)[lb]{\smash{$b_e$}}}%
    \put(0.39269511,0.31269069){\color[rgb]{0,0,0}\makebox(0,0)[lb]{\smash{$b_f$}}}%
    \put(0.3936589,0.37193515){\color[rgb]{0,0,0}\makebox(0,0)[lb]{\smash{$b_g$}}}%
    \put(0.39376667,0.19167427){\color[rgb]{0,0,0}\makebox(0,0)[lb]{\smash{$b_0$}}}%
    \put(0.82700707,0.01580858){\color[rgb]{0,0,0}\makebox(0,0)[b]{ \smash{\bf \shortstack[c]{Multiscale\\Hierarchy of\\Particle\\Filters}}}}%
    \put(0.02236033,0.04757514){\color[rgb]{0,0,0}\makebox(0,0)[lb]{\smash{\bf Trajectories}}}%
    \put(0.33639107,0.01521698){\color[rgb]{0,0,0}\makebox(0,0)[b]{\smash{\bf \shortstack[c]{Filtration\\of\\Trajectory\\Abstractions}}}}%
    \put(0.5889532,0.02630858){\color[rgb]{0,0,0}\makebox(0,0)[b]{\smash{\bf \shortstack[c]{Hierarchical\\Class\\Tree}}}}%
    \put(0.14230975,0.06466637){\color[rgb]{0,0,0}\makebox(0,0)[lb]{\smash{\emph{\shortstack[c]{Hierarchical\\Agglomerative\\Clustering}}}}}%
    \put(0.4190305,0.06246964){\color[rgb]{0,0,0}\makebox(0,0)[lb]{\smash{\emph{\shortstack[c]{Dynamics\\Learning}}}}}%
    \put(0.68477022,0.06258289){\color[rgb]{0,0,0}\makebox(0,0)[lb]{\smash{\emph{Prior}}}}%
    \put(0.91842793,0.06898185){\color[rgb]{0,0,0}\makebox(0,0)[lb]{\smash{\bf \shortstack[c]{Mixed\\Observations}}}}%
  \end{picture}%
}
\endgroup%